\documentclass[10pt,twocolumn,letterpaper]{article}

\usepackage[pagenumbers]{cvpr} 

\usepackage{graphicx}
\usepackage{amsmath}
\usepackage{amssymb}
\usepackage{booktabs}
\usepackage{multirow}
\usepackage{diagbox}
\usepackage{color}
\usepackage[pagebackref,breaklinks,colorlinks]{hyperref}

\usepackage[capitalize]{cleveref}
\crefname{section}{Sec.}{Secs.}
\Crefname{section}{Section}{Sections}
\Crefname{table}{Table}{Tables}
\crefname{table}{Tab.}{Tabs.}

\def\cvprPaperID{4102} 
\def\confName{CVPR}
\def\confYear{2022}

\begin{document}

\title{Iterative Deep Homography Estimation}

\author{Si-Yuan Cao, Jianxin Hu, Zehua Sheng, Hui-Liang Shen \\
Zhejiang University\\
{\tt\small karlcao@hotmail.com}, {\tt\small \{hujianxin,shengzehua,shenhl\}@zju.edu.cn}
}

\maketitle

\begin{abstract}
We propose Iterative Homography Network, namely IHN, a new deep homography estimation architecture. Different from previous works that achieve iterative refinement by network cascading or untrainable IC-LK iterator, the iterator of IHN has tied weights and is completely trainable. IHN achieves state-of-the-art accuracy on several datasets including challenging scenes. We propose 2 versions of IHN: (1) IHN for static scenes, (2) IHN-mov for dynamic scenes with moving objects. Both versions can be arranged in 1-scale for efficiency or 2-scale for accuracy. We show that the basic 1-scale IHN already outperforms most of the existing methods. On a variety of datasets, the 2-scale IHN outperforms all competitors by a large gap. We introduce IHN-mov by producing an inlier mask to further improve the estimation accuracy of moving-objects scenes. We experimentally show that the iterative framework of IHN can achieve 95\% error reduction while considerably saving network parameters. When processing sequential image pairs, IHN can achieve 32.7 fps, which is about 8$\times$ the speed of IC-LK iterator. Source code is available at \url{https://github.com/imdumpl78/IHN}.
\end{abstract}

\section{Introduction}
\label{sec:intro}

Homography estimation aims to find the global perspective transform between two images. It serves as a crucial step in a widely range of computer vision tasks such as image/video stitching \cite{szeliski2006image,guo2016joint}, video stabilization \cite{liu2013bundled}, SLAM \cite{engel2014lsd,mur2015orb}, augmented reality \cite{simon2000markerless}, GPS denied navigation \cite{goforth2019gps, zhao2021deep}, and multimodal image fusion \cite{zhou2019integrated, ying2021unaligned}. 

The approaches in the literature can be roughly categorized into photometric-based and feature-based ones \cite{szeliski2006image}. Photometric-based approaches aim to estimate homography from pixel intensities. The Lucas-Kanade (LK) algorithm \cite{lucas1981iterative,baker2004lucas} is the most widely adopted photometric-based approach, which iteratively estimates the residual homography using a pre-computed iterator. Feature-based approaches usually consist of three steps: feature extraction, feature matching, and homography estimation \cite{szeliski2006image}. The well-known feature extractors are SIFT \cite{lowe2004distinctive}, SURF \cite{bay2006surf}, and ORB \cite{bay2006surf}. Homography estimation methods include RANSAC \cite{fischler1981random}, DLT \cite{dubrofsky2009homography}, and MAGSAC\cite{barath2019magsac}. 

\begin{figure}
  \centering
  \begin{subfigure}{1\linewidth}
  \centering
  \includegraphics[scale=0.125]{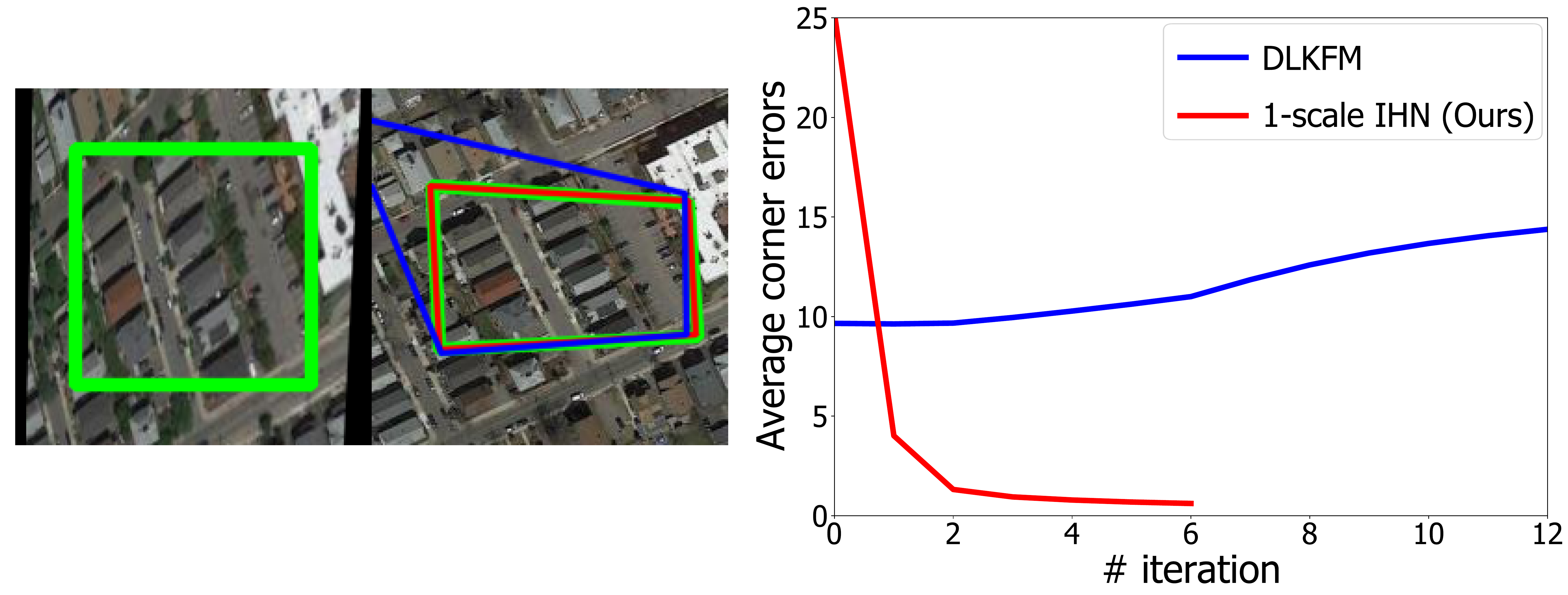}
    \caption{Homography estimation on Google Earth.}
    \label{fig:googleearth}
  \end{subfigure}
  \hfill
  \begin{subfigure}{1\linewidth}
  \centering
  \includegraphics[scale=0.125]{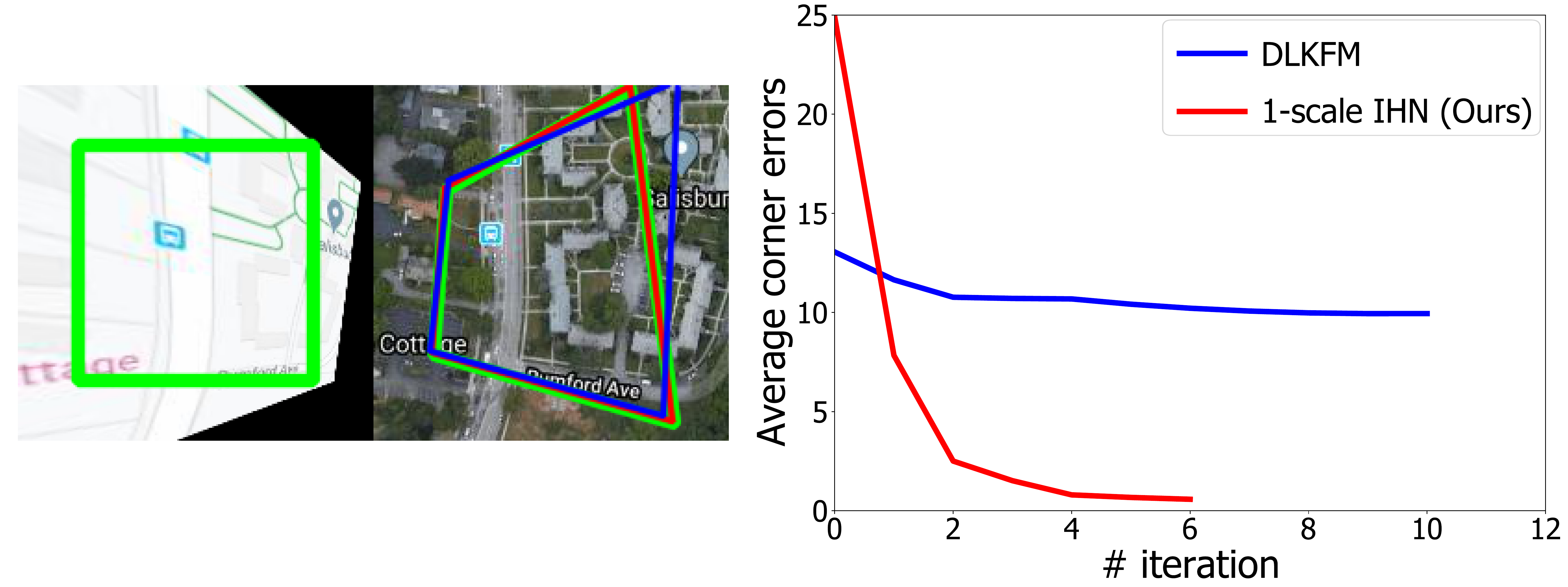}
  \caption{Homography estimation on Google Map \& Satellite.}
  \label{fig:googlemap}
  \end{subfigure}  
  \hfill
   \vspace{-3mm}
  \caption{Visualization of homography estimation with average corner error (ACE) at each iteration of our IHN and DLKFM \cite{zhao2021deep} (which use the traditional IC-LK iterator). Left 2 images: image pair for homography estimation with the source image $I_{\mathrm{S}}$ on the left and the target image $I_{\mathrm{T}}$ on the right. Green polygons denote the ground-truth position of $I_{\mathrm{S}}$ on $I_{\mathrm{T}}$. Blue polygons denote the estimated position using MHN+DLKFM. Red polygons denote the estimated position using our IHN. Right plot: ACEs during first 12 iterations. IHN stops at iteration 6 while DLKFM has a dynamic stop criterion which iterates 21 times averagely.}
   \vspace{-5mm}
  \label{fig:google}
\end{figure}

Recently, deep homography methods have attracted increasing interest due to their superior performance. The VGG-style network, first proposed by DeTone \etal. \cite{detone2016deep}, is adopted to directly estimate homography. Based on this work, many recent methods have been introduced to boost the estimation accuracy by cascading multiple VGG-style networks \cite{erlik2017homography,zhou2019stn,le2020deep}. The cascading is actually a way of iteration that can significantly improve the estimation performance. However, this way of iteration is limited to a fixed number of cascading, and more network cascading does not necessarily lead to better performance \cite{le2020deep}. 

To further improve the homography estimation accuracy, some works \cite{chang2017clkn,zhao2021deep} take the Lucas-Kanade (LK) algorithm as an untrainable iterator and combine it with CNNs. However, the approximation of the Hessian matrix in the LK algorithm fails when the Jacobian matrix is rank-deficient \cite{nocedal2006numerical}. Even worse, the training of the network is limited to the feature extractor, which means the above drawback is theoretically unavoidable. 

To cope with these issues, we propose iterative homography network (IHN), which is completely trainable. We introduce 2 versions of IHN: (1) IHN for static scenes, (2) IHN-mov for dynamic scenes with moving objects. Both versions can be arranged in 1-scale for efficiency or 2-scale for accuracy. We show that the basic 1-scale IHN already outperforms most of the existing methods. On MSCOCO\cite{lin2014microsoft}, the 2-scale IHN outperforms all competitors by a large gap, with more than 90\% of the average corner errors (ACEs) lower than 0.1 pixels. On cross-modal datasets \cite{zhao2021deep}, IHN outperforms the state-of-the-art method MHN+DLKFM \cite{zhao2021deep} that uses the traditional LK iterator in refining the deep homography estimation of MHN \cite{le2020deep}. On moving-objects data\cite{wang2016spid}, IHN and IHN-mov both surpass the competitors. IHN-mov gains further accuracy improvement by producing an inlier mask mimicking the essence of RANSAC \cite{fischler1981random}.

The motivation of IHN comes from the traditional IC-LK iterator \cite{baker2004lucas}. Different from IC-LK, IHN is completely trainable and thus can learn prior information for the residual homography prediction directly from data. Fig. \ref{fig:google} illustrates the homography estimation results and average corner errors (ACEs) at each iteration for our IHN and DLKFM \cite{zhao2021deep} that use the IC-LK iterator. The initial ACEs at iteration 0 differs because DLKFM uses the preliminary estimation from MHN. It is observed that our basic 1-scale IHN can produce an accurate homography estimation within 6 iterations, while DLKFM using the traditional IC-LK fails. Furthermore, IHN can process image pairs in sequence at 32.7 fps, which is about 8$\times$ the speed of the IC-LK iterator. We further explore the effectiveness of our iterative framework by replacing our global motion aggregator (GMA, which directly estimate the residual homography) with the network architectures in \cite{detone2016deep,le2020deep,zhang2020content}. Experimental results show that compared to feature/image concatenation strategy in \cite{detone2016deep,le2020deep,zhang2020content,zhou2019stn,nguyen2018unsupervised}, our iterative framework achieves about 95\% error reduction. Thanks to the iterative framework, our global motion aggregator can achieve comparable accuracy with much fewer parameters than previous architectures, \emph{e.g.}, 36.4\% of the architecture in \cite{le2020deep} and 3.6\% in \cite{zhang2020content}. The same deep iterative concept that significantly improves optical flow estimation accuracy has been proposed in RAFT \cite{teed2020raft}, which inspires us to construct a completely trainable deep homography estimation network.

To summarize, the main contributions of this work are as follows:
\begin{itemize}
\item We propose an iterative homography network, namely IHN, which is completely trainable. IHN achieves state-of-the-art accuracy on several datasets including challenging scenes. The iteration of IHN is stable and doesn't require extra parameters. 
\item We show that the proposed iterative framework is vital for an accurate homography estimation, which can achieve 95\% error reduction regardless of a specific network architecture design. The iterative framework also enables considerable parameter saving.
\item We specially design a network architecture IHN-mov for the moving-objects scenes where homography assumption is violated. The network produces an inlier mask mimicking the essence of RANSAC that can further benefit the homography estimation.
\end{itemize}

\section{Related Work}
\label{sec:rela}
We make a brief introduction of deep homography estimation, challenges in homography estimation, and iterative homography estimation that are most relevant to our method. For the basic knowledge of homography estimation, the readers are referred to \cite{szeliski2006image,zitova2003image}.

\textbf{Deep Homography Estimation.} Deep homography estimation is first proposed by DeTone \etal \cite{detone2016deep}, who adopted a VGG-style network to directly predict the homography between the concatenated source and target images. Following this pioneering work, several works \cite{erlik2017homography,zhou2019stn,le2020deep} proposed to cascade multiple VGG-style networks to improve the homography estimation accuracy. Nowruzi \etal \cite{erlik2017homography} proposed to arrange similar stacked networks to successively refine the homography estimation. Le \etal \cite{le2020deep} proposed to use multiscale VGG-style networks to iteratively estimate the residual homography. Nevertheless, compared to the Lucas-Kanade iterator \cite{lucas1981iterative, baker2004lucas}, the cascaded deep homography methods still lack accuracy \cite{chang2017clkn,zhao2021deep}. 

Another kind of works accomplishes iterative deep homography estimation by combining the LK algorithm with CNNs. Chang \etal \cite{chang2017clkn} adopted the inverse compositional LK (IC-LK) iterator as an untrainable layer of the deep network. A CNN is employed to extract the feature maps that are optimal for the IC-LK iterator. Zhao \etal \cite{zhao2021deep} proposed to construct a one-channel deep Lucas-Kanade feature map (DLKFM) using CNNs. The DLKFM is then sent into the IC-LK iterator. Similar approaches can be found in \cite{wang2018deep, goforth2019gps}. However, the LK iterator is untrainable, and therefore theoretically the drawbacks such as rank-deficient Jacobian cannot be avoided.
 
\begin{figure*}[t]
  \centering
  \includegraphics[scale=0.25]{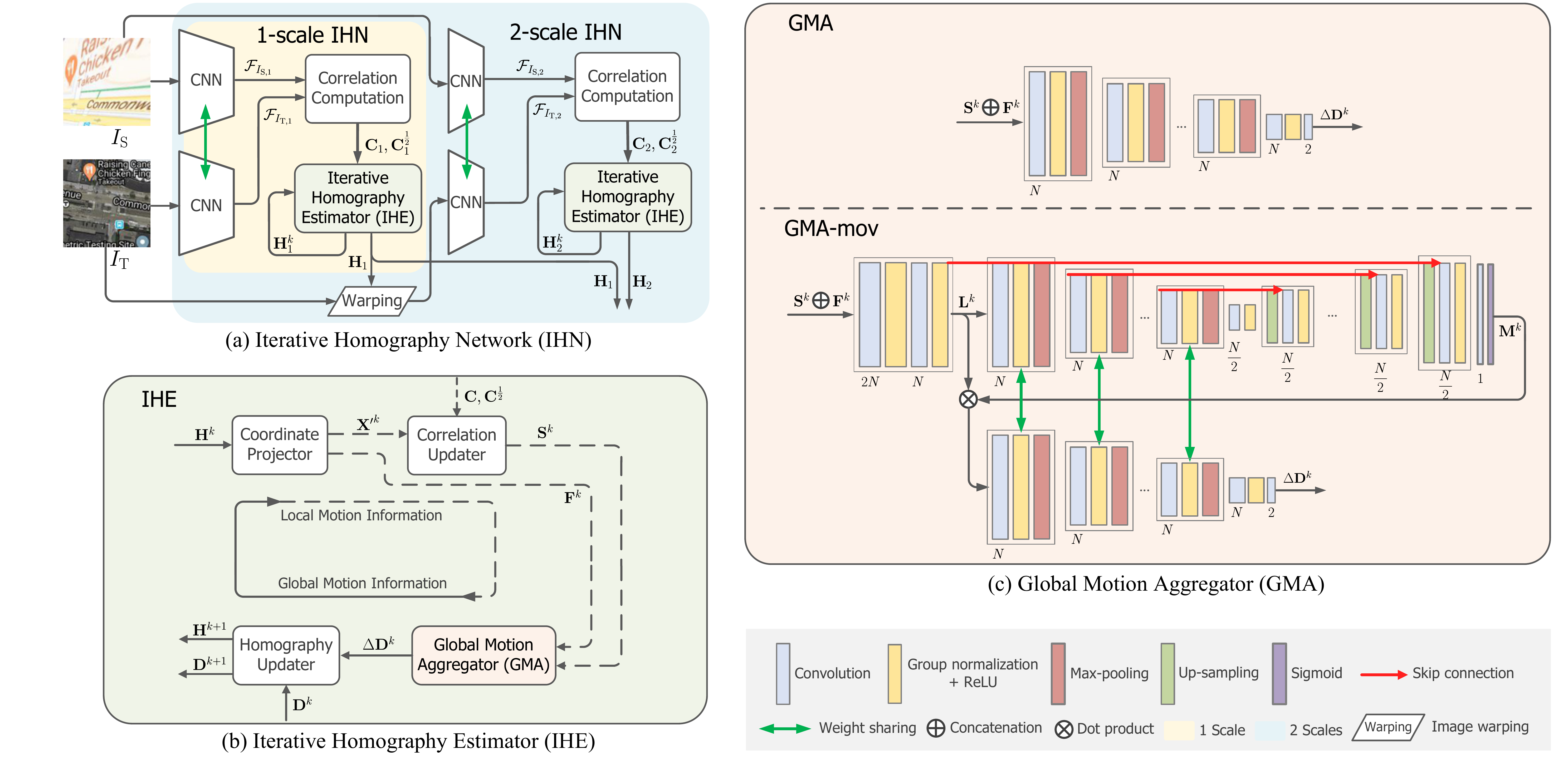}
   \vspace{-3mm}
   \caption{Schematics and detailed architectures of iterative homography network (IHN). (a) Overall schematic of IHN, including the illustration of basic 1-scale IHN and 2-scale IHN. (b) Structure of the iterative homography estimator (IHE), which plays a leading role of realizing the iterative homography refinement. (c) Architecture of the global motion aggregator (GMA) that directly estimates residual homography with GMA designed for the static scenes and GMA-mov for the moving-objects scenes. Please refer to the text for details.}
    \vspace{-5mm}
   \label{fig:IHN}
\end{figure*}

\textbf{Challenges in Homography Estimation.} According to the recent works \cite{shao2021localtrans,zhao2021deep,le2020deep,nguyen2018unsupervised,zhang2020content}, there are mainly two challenges in homography estimation. The first challenge comes from the photometric inconsistency such as illumination change or modality variation. Nguyen \etal \cite{nguyen2018unsupervised} proposed an unsupervised learning method to improve the network capacity on illumination variation. Zhao \etal \cite{zhao2021deep} proposed to extract intensity consistent DLKFM for cross-modal inputs. They estimated the homography using multiple cascaded VGG-style networks MHN \cite{le2020deep} and refined it using the IC-LK iterator. The second challenge comes from the violation of the homography assumption. For example, in scenes with moving objects, the matching between the source and target images does not always satisfy a uniform homography. Zhang \etal \cite{zhang2020content} proposed a mask predictor network that weights the feature maps to achieve content-aware homography estimation. However, as the masks are separately computed on the source and target images, the method cannot reject matching outliers. Le \etal \cite{le2020deep} proposed to teach the network a moving object mask produced by the PWC-Net \cite{sun2018pwc} optical flow. However, the mask cannot be learned if the optical flow estimation fails.

\textbf{Iterative Homography Estimation.} The most widely used iterative homography estimation framework is the Lucas-Kanade (LK) algorithm \cite{lucas1981iterative,baker2004lucas}. The objective function of the most widely adopted inverse compositional Lucas-Kanade (IC-LK) algorithm is
\begin{equation}
\min_{\Delta \mathbf{H}} \|I_{\mathrm{T}}(W(\mathbf{X};\Delta \mathbf{H}) - I_{\mathrm{S}}(W(\mathbf{X};\mathbf{H}))\|^2_2,
  \label{eq:ICLK_obj}
\end{equation}
where $\Delta \mathbf{H}$ denotes the residual homography matrix, and $W$ denote the coordinate warping operation. By conducting a first-order Taylor expansion on $I_{\mathrm{T}}(W(\mathbf{X};\Delta \mathbf{H})$, a closed-form solution of $\Delta \mathbf{H}$ can be computed as 
\begin{equation}
\Delta \mathbf{H} = (\mathbf{J}^\mathsf{T}\mathbf{J})^{-1}\mathbf{J}^\mathsf{T}\mathbf{r},
  \label{eq:ICLK_iter}
\end{equation}
where $\mathbf{J}$ denotes the Jacobian matrix of $I_{\mathrm{T}}$, $\mathbf{J}^\mathsf{T}\mathbf{J}$ is the approximation of the Hessian matrix. $\mathbf{r}=\mathrm{vec}(I_{\mathrm{T}} - I_{\mathrm{S}}(W(\mathbf{X};\mathbf{H})))$ represents the vectorized residual image that is updated in every iteration. 

The IC-LK algorithm iteratively estimates homography in 3 steps: (1) updating $I_{\mathrm{S}}(W(\mathbf{X};\mathbf{H}))$ using local coordinates, and thus updating local information in $\mathbf{r}$; (2) aggregating residual global homography using Eq. (\ref{eq:ICLK_iter}); (3) projecting updated global homography into local coordinates for updating local information in the next iteration.

To the best of our knowledge, there isn't any completely iterative trainable deep homography estimation network before our IHN. The most related work is the deep iterative optical flow estimation proposed in RAFT \cite{teed2020raft}. 

\section{Method}
\label{sec:method}

The overview of our iterative homography network (IHN) is illustrated in Fig. \ref{fig:IHN}a. IHN takes in a pair of source image $I_\mathrm{S}$ and target image $I_\mathrm{T}$ and outputs the estimated homography matrix $\mathbf{H}_1$ (1 scale) or $\mathbf{H}_1$ and $\mathbf{H}_2$ (2 scales). The main steps of IHN include feature extraction using CNN, correlation computation, and recurrent homography estimation using an iterative homography estimator (IHE). 
\subsection{Feature Extraction}
\label{feature_extraction}

We extract the feature maps of the source and target images using a Siamese CNN. We take the combination of $1$ max-pooling layer (stride $2$) and $2$ residual blocks as a basic unit. The images are first processed by $1$ convolutional block with kernel size $7 \times 7$. We then add $q$ basic units to produce the $1/2^{q} \times 1/2^{q}$ resolution feature maps. The feature maps are finally reprojected by $1$ linear convolutional layer with kernel $1 \times 1$. Specifically, we set $q=2$ in practical implementation. As illustrated in Fig. \ref{fig:IHN}a, the 1-scale IHN uses the feature maps at $1/4 \times 1/4$ resolution, and the 2-scale IHN uses both feature maps at $1/4 \times 1/4$ and $1/2 \times 1/2$ resolution. For the 2-scale IHN, the first basic units of both resolution feature maps share the same weights. We found that the 1-scale IHN already achieves considerable homography estimation performance, while the 2-scale IHN can further improve the accuracy. The Siamese CNN is also used for cross-modal data. We will show in Section \ref{subsec:cross-modal_exp} that our IHN can produce promising homography estimation without a specific feature extractor setting (\eg, the pseudo-Siamese network in \cite{zhao2021deep} for cross-modal inputs). 
\subsection{Correlation}
\label{subsec:corr}

Different from most previous deep homography estimation works \cite{detone2016deep,erlik2017homography,le2020deep,zhou2019stn,nguyen2018unsupervised}, we explicitly compute the correlation to enable iterative refinement. Let us denote the feature maps of the source and target images as $\mathcal{F}_{I_{\mathrm{S}}}\in \mathbb{R}^{D \times H\times W }$ and $\mathcal{F}_{I_{\mathrm{T}}}\in \mathbb{R}^{D \times H\times W}$. We set $D=256$ for all feature maps. We compute the pairwise correlation, namely the correlation volume as
\begin{equation}
  \mathbf{C}(\mathbf{x}_{\mathrm{S}},\mathbf{x}_{\mathrm{T}}) = \mathrm{ReLU}(\mathcal{F}_{I_{\mathrm{S}}}(\mathbf{x}_{\mathrm{S}})^{\mathsf{T}} \mathcal{F}_{I_{\mathrm{T}}}(\mathbf{x}_{\mathrm{T}})),
  \label{eq:correlation_volume}
\end{equation}
where $\mathbf{x}_{\mathrm{S}}$ and $\mathbf{x}_{\mathrm{T}}$ denote the coordinate position of the source and target feature maps. The correlation volume $\mathbf{C}$ is of size $H \times W \times H \times W$. In one iteration, a fixed search window is sampled from $\mathbf{C}$ by the correlation updater, which is described in detail in Section \ref{subsec:ihe}. As mentioned in \cite{teed2020raft}, the correlation volume can also be computed on demand during the iteration, which can reduce the space complexity.  

\textbf{Correlation Pooling.} To enlarge the perception range within a feature scale, we conduct average pooling on $\mathbf{C}$ at stride 2 at the last 2 dimensions to form another correlation volume $\mathbf{C}^{\frac{1}{2}} $, which is of size $ H \times W \times H/2 \times W/2$. For both volumes, we use search windows of the same size, which means that the sampling operation on $\mathbf{C}^{\frac{1}{2}}$ has a $2\times 2$ larger perception range compared to $\mathbf{C}$. 

\subsection{Iterative Homography Estimator}
\label{subsec:ihe}

We design our iterative homography estimator (IHE) under the inspiration of the IC-LK iterator. IHE plays a leading role in the realization of the iterative homography refinement. As illustrated in Fig. \ref{fig:IHN}a and \ref{fig:IHN}b, IHE takes in the correlation volume $\mathbf{C},\mathbf{C}^{\frac{1}{2}}$ and outputs the estimated homography $\mathbf{H}$. From the coordinate projector to the global motion aggregator, local motion information is aggregated into global homography estimation. From the global motion aggregator back to the coordinate projector (in the next iteration), global homography estimation is converted to local coordinates for local information update. Similar essence can be found in the IC-LK iterator.

\begin{figure}[t]
  \centering
  \includegraphics[scale=0.55]{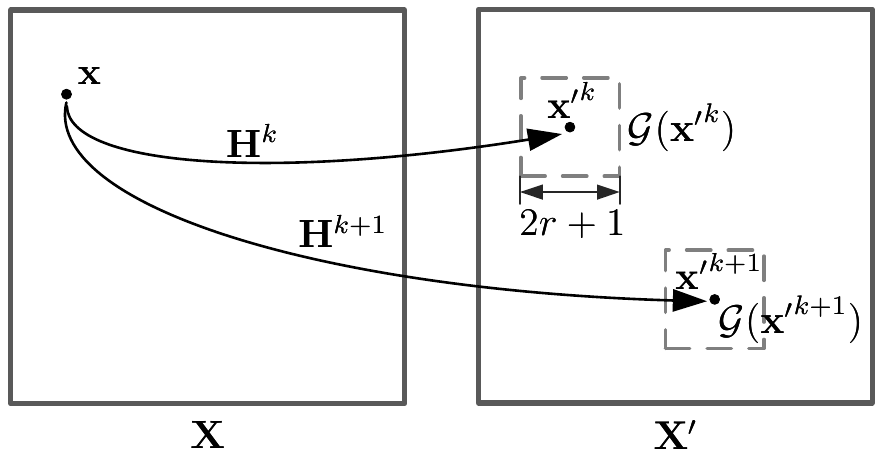}
   \vspace{-3mm}
   \caption{Illustration of the iterative process of coordinate projector and correlation updater. Left: coordinate of $\mathcal{F}_{I_\mathrm{S}}$, denoted as $\mathbf{X}$. Right: coordinate of $\mathcal{F}_{I_\mathrm{T}}$, denoted as $\mathbf{X'}$.}
    \vspace{-6mm}
   \label{fig:coordinate_projector}
\end{figure}

\textbf{Coordinate Projector.} Taking iteration $k$ for example, the point-wise correspondence between the source feature map $\mathcal{F}_{I_{\mathrm{S}}}$ and the target feature map $\mathcal{F}_{I_{\mathrm{T}}}$ is mapped by the present homography matrix $\mathbf{H}^{k}$. Let's denote $\mathbf{X}$ as the meshgrid coordinate set of $\mathcal{F}_{I_{\mathrm{S}}}$, and its corresponding meshgrid coordinate set in $\mathcal{F}_{I_{\mathrm{T}}}$ as $\mathbf{X'}$. For each coordinate position, we denote $\mathbf{x}=(u,v), \mathbf{x} \in \mathbf{X}$ and $\mathbf{x'}=(u',v'), \mathbf{x'} \in \mathbf{X'}$. The point-wise correspondence of $\mathbf{X}$ and $\mathbf{X'}^k$ is projected by $\mathbf{H}^{k}$ using
\begin{equation}
\begin{bmatrix} u'^k \\ v'^k \\ 1 \end{bmatrix} \sim \begin{bmatrix} \mathbf{H}_{11}^k & \mathbf{H}_{12}^k & \mathbf{H}_{13}^k \\ \mathbf{H}_{21}^k & \mathbf{H}_{22}^k & \mathbf{H}_{23}^k \\ \mathbf{H}_{31}^k & \mathbf{H}_{32}^k & 1 \end{bmatrix} \begin{bmatrix} u \\ v \\ 1 \end{bmatrix}.
  \label{eq:homography_warp_coordinates}
\end{equation}
 We illustrate the coordinate projection in iteration $k$ and $k+1$ in Fig. \ref{fig:coordinate_projector}. To further facilitate the learning of local motion information, we also compute the homography flow $\mathbf{F}^{k}$ as
\begin{equation}
  \mathbf{F}^{k} = \mathbf{X'}^{k} - \mathbf{X}.
  \label{eq:homography_flow}
\end{equation}
$\mathbf{F}^{k}$ is then sent to the global motion aggregator.

\textbf{Correlation Updater.} The correlation updater samples the correlation volume $\mathbf{C}$ using homography projected coordinate $\mathbf{X'}^{k}$ and outputs updated correlation slice $\mathbf{S}^{k}$. The sampling process can be expressed as
\begin{equation}
  \mathbf{S}^k(\mathbf{x}) = \mathbf{C}(\mathbf{x}, \mathcal{G}_r(\mathbf{x'}^{k})),
  \label{eq:local_correlation_slice}
\end{equation}
where $\mathcal{G}_r(\mathbf{x'}^{k})$ denotes a local square grid of fixed search radius $r$. The square grid sampling is illustrated in Fig. \ref{fig:coordinate_projector}. Note that the correlation slice is also sampled on the pooled correlation volume $\mathbf{C}^{\frac{1}{2}}$ to make $\mathbf{S}^{\frac{1}{2},k}$ that have $2 \times 2$ larger perception range. 

\textbf{Global Motion Aggregator.} The residual homography is estimated by the global motion aggregator, with the homography matrix parameterized by the displacement vectors of the 4 corner points as in \cite{detone2016deep,le2020deep,shao2021localtrans}. 

As illustrated in Fig. \ref{fig:IHN}c, we design the GMA for static scenes and the GMA-mov for moving-objects scenes. $N$ denotes the number of filters of convolutional kernels.

GMA is mainly composed of multiple basic units. Each basic unit includes a $3 \times 3$ convolutional block, $1$ group normalization \cite{wu2018group} + ReLU \cite{maas2013rectifier}, and $1$ max-pooling layer (stride $2$). We continually add the basic unit until the spatial resolution of the feature map reaches $2 \times 2$. Then a convolutional block projects the feature map into a $2 \times 2 \times 2$ cube $\Delta \mathbf{D}^{k}$, which is the estimated residual displacement vectors of the 4 corner points. In iteration $k$, GMA takes the concatenated correlation slice $\mathbf{S}^k$ and homography flow $\mathbf{F}^k$ as input. 

GMA-mov is specifically designed for scenes with moving objects. GMA-mov explicitly produces a mask $\mathbf{M}^k$ to weight the matching inliers that meet the homography transform assumption. It is worth noting that, different from \cite{le2020deep} that needs additional optical flow supervision or \cite{zhang2020content} that relies on image content, our GMA-mov can produce an inlier mask based on the combined local and global motion information as RANSAC. The mask can be generated without supervision, yet can improve the homography estimation accuracy. As illustrated in Fig. \ref{fig:IHN}c, GMA-mov explicitly encodes the point-wise local motion information into feature map $\mathbf{L}^k$ using $1$ convolutional block. $\mathbf{L}^k$ is then sent into multiple basic units as in GMA to preliminarily extract the $N \times 2 \times 2$ feature map containing the global motion information. Different from GMA, the extracted global motion information is not directly used for the residual homography estimation, but the inlier mask prediction. The latter half part of GMA-mov progressively upsamples the feature map containing the global motion information and combines it with the local motion information by skip connections. The inlier mask $\mathbf{M}^k$ with the same size as $\mathbf{L}^k$ is predicted by a sigmoid function. $\mathbf{M}^k$ and $\mathbf{L}^k$ are taken dot product and sent into the same structure as GMA to produce the residual homography estimation. We note that the basic units used for extracting the preliminary global motion information and residual homography estimation share the same weights.

\textbf{Homography Updater.} As in \cite{detone2016deep,le2020deep,shao2021localtrans}, we parameterize the homography matrix using the displacement vectors of the 4 corner points of an image, namely the displacement cube $\mathbf{D}$. In iteration $k$, $\mathbf{D}$ is updated as
\begin{equation}
  \mathbf{D}^{k+1} = \mathbf{D}^{k} + \Delta \mathbf{D}^{k}.
  \label{eq:displacement_vectors_update}
\end{equation}
With $\mathbf{D}^{k+1}$, it is convenient to get the homography matrix $\mathbf{H}^{k+1}$ by the least square method, the direct linear transform \cite{abdel2015direct} or other methods. The updated $\mathbf{H}^{k+1}$ will be sent into the coordinate projector in the next iteration. The initial displacement cube is set to $\mathbf{D}^0 = \mathbf{0}$, which means identical transform $\mathbf{H}$. 
\subsection{Multiscale Strategy} 

We introduce a multiscale strategy that can further improve the homography estimation accuracy. We note that, according to our experiments, the 1-scale IHN can outperform most existing homography estimation methods. As illustrated in Fig. \ref{fig:IHN}a, another scale of IHN with its correlation volume computed on the $1/2 \times 1/2$ resolution feature maps is attached to the 1-scale IHN. The target image $I_\mathrm{T}$ is warped using the estimated homography $\mathbf{H}_1$ at $1/4 \times 1/4$ resolution. The bottom right subscript, namely $1$ or $2$, denotes the computed result at the $1/4 \times 1/4$ resolution or the $1/2 \times 1/2$ resolution. The homography matrices $\mathbf{H}_1$ and $\mathbf{H}_2$ from the 2 scales are composed to produce the final homography estimation for the 2-scale IHN as in \cite{le2020deep,shao2021localtrans}. In Section \ref{sec:experiments}, we show that IHN only needs $2$ scales to achieve very high accuracy than the 3 or 4 scales IC-LK based methods \cite{chang2017clkn,zhao2021deep}.
\subsection{Loss Function} 

We apply supervision on the $L_1$ distance between the ground-truth displacement $\mathbf{D}_{\mathrm{gt}}$ and the estimated displacement $\mathbf{D}$ at each iteration. A weighted sum of all the iteration is computed as the loss function
\begin{equation}
  L = \sum_{k=0}^{K-1} \alpha^{(K-k-1)}|\mathbf{D}^{k+1} - \mathbf{D}_{\mathrm{gt}}|,
  \label{eq:supervision}
\end{equation}
where $K$ denotes the total iteration at one resolution, and $k$ ranges from $0$ to $K-1$. If the multiscale strategy is applied, the loss of both resolutions is separately computed and summed to make the final loss.

\begin{figure*}
  \centering
  \begin{subfigure}{0.495\linewidth}
  \centering
  \includegraphics[scale=0.3]{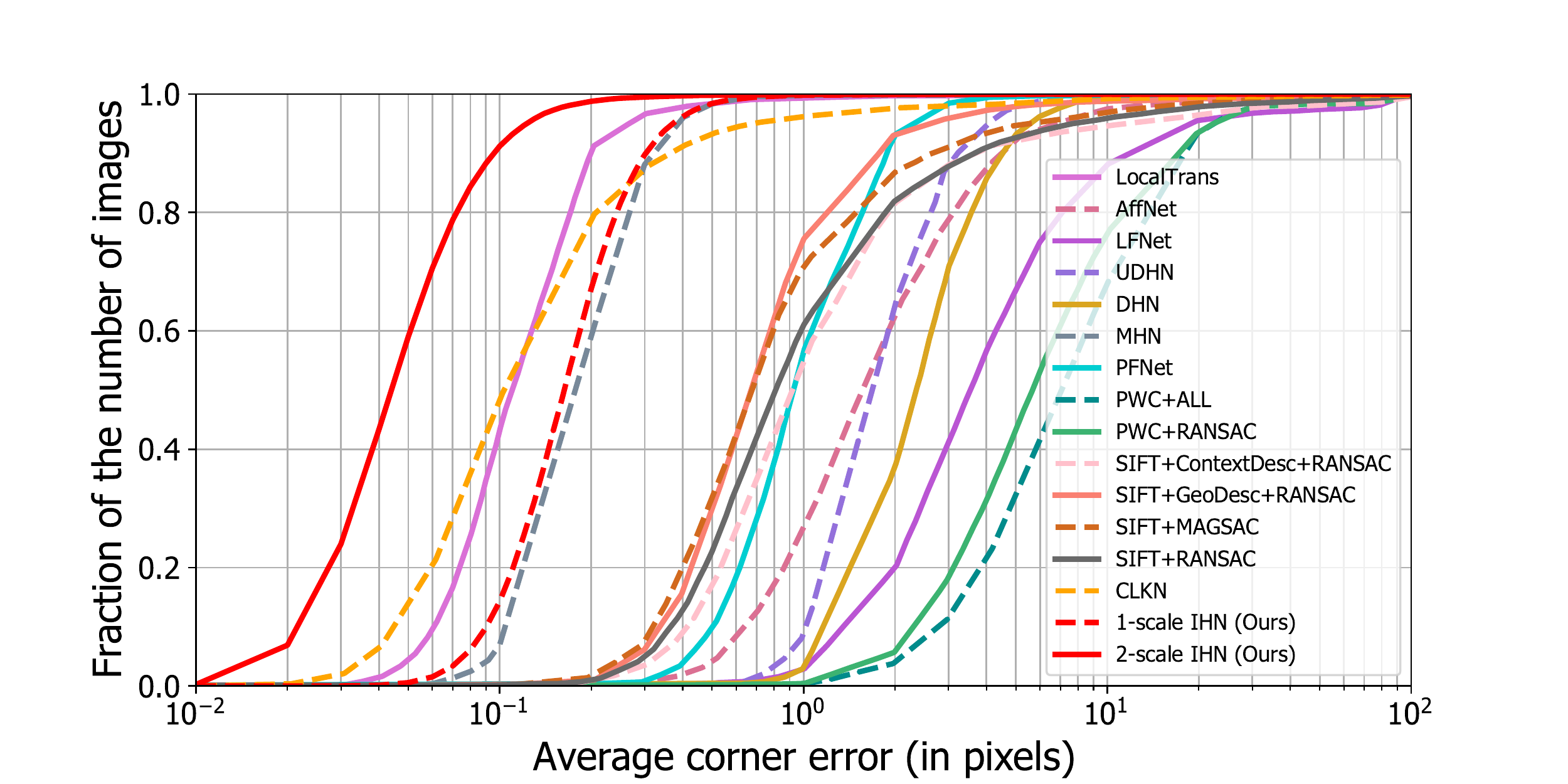}
  \vspace{-1mm}
    \caption{Evaluation on MSCOCO.}
    \vspace{-1mm}
    \label{fig:eva_mscoco}
  \end{subfigure}
  \hfill
  \begin{subfigure}{0.50\linewidth}
  \centering
  \includegraphics[scale=0.3]{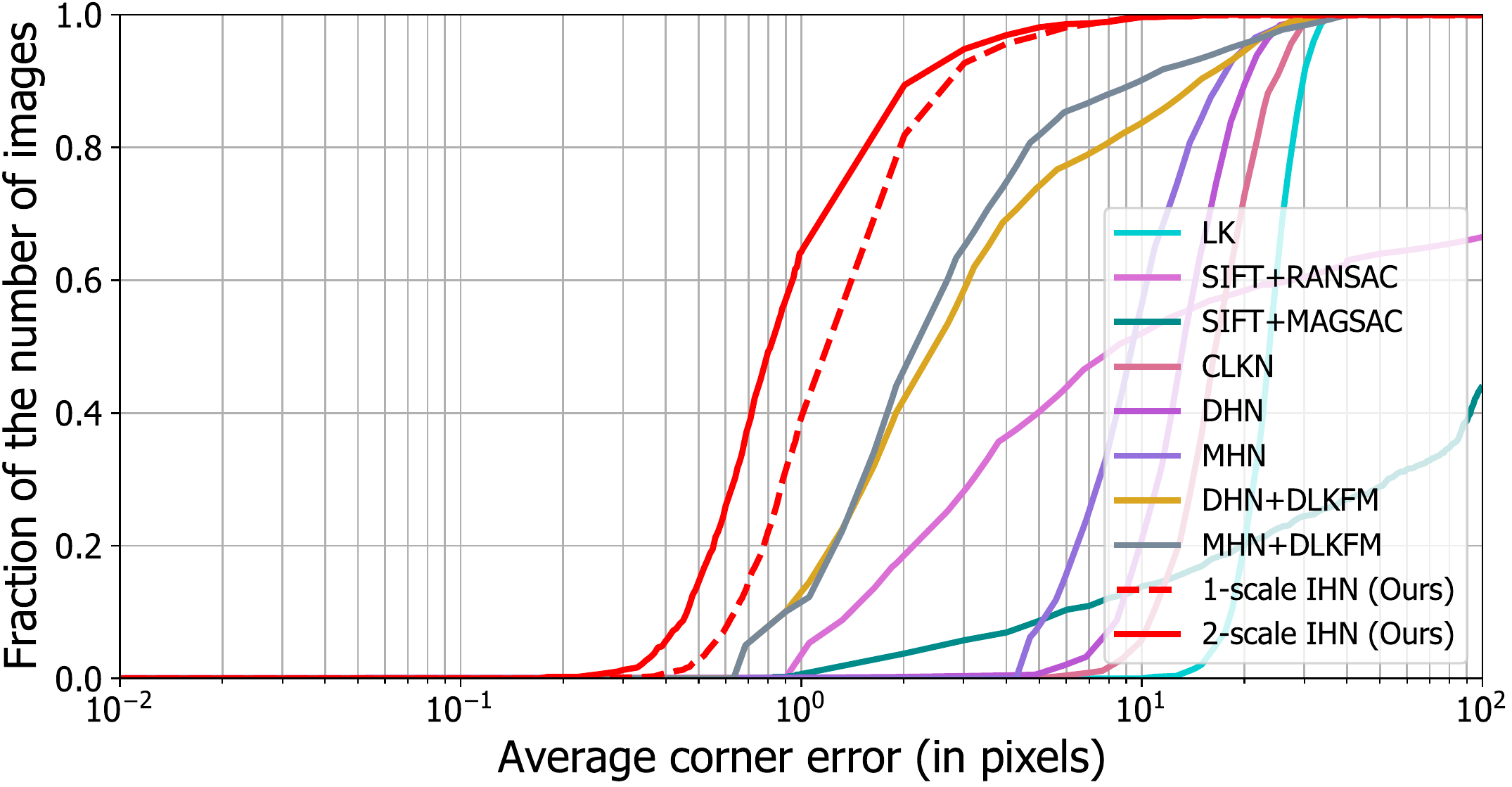}
  \vspace{-1mm}
  \caption{Evaluation on Google Earth.}
  \vspace{-1mm}
  \label{fig:eva_ggearth}
  \end{subfigure}  
  \hfill
  \begin{subfigure}{0.495\linewidth}
  \centering
  \includegraphics[scale=0.3]{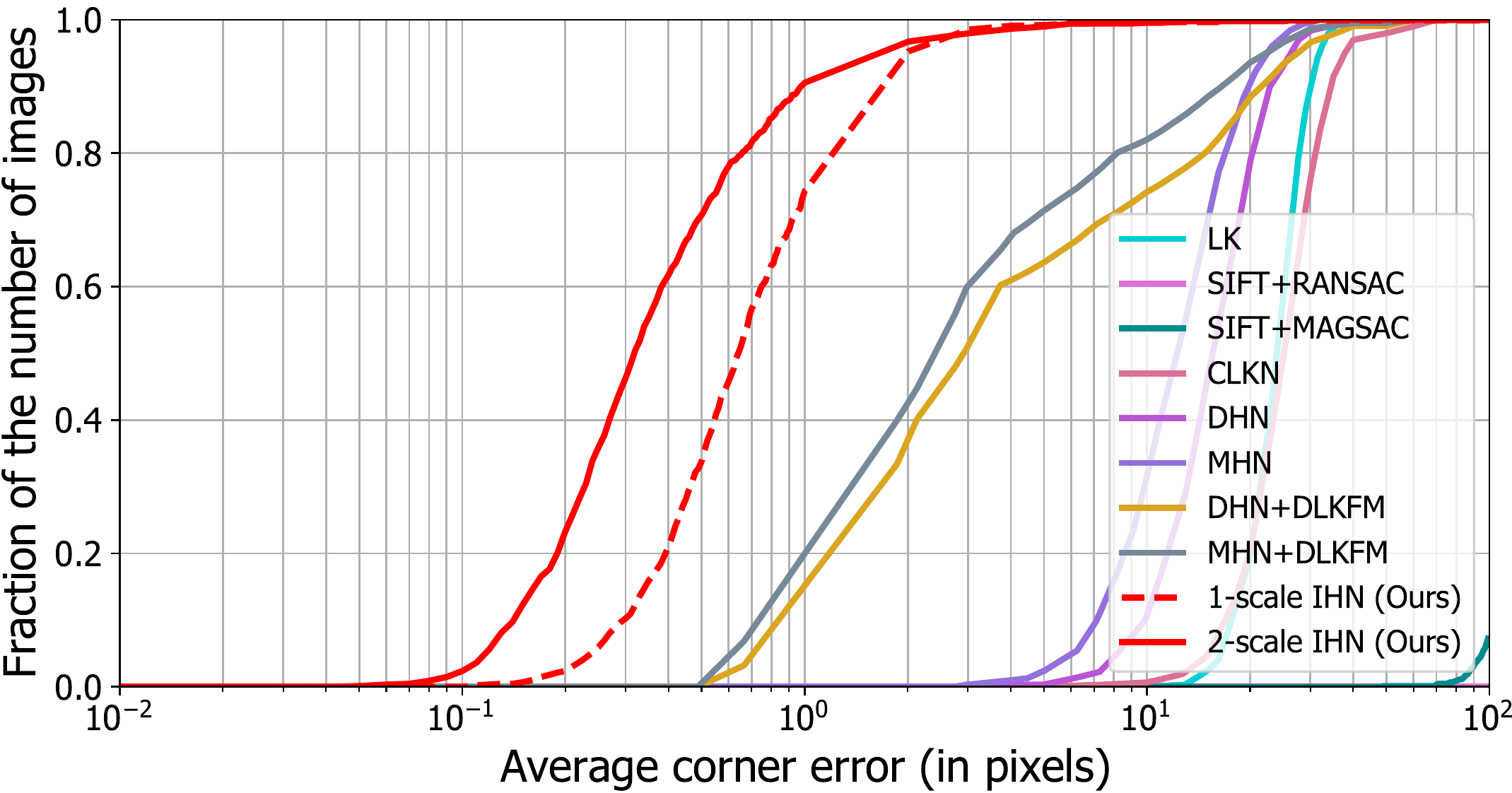}
  \vspace{-1mm}
  \caption{Evaluation on Google Map \& Satellite.}
  \vspace{-1mm}
  \label{fig:eva_ggmap}
  \end{subfigure} 
  \hfill
  \begin{subfigure}{0.495\linewidth}
  \centering
  \includegraphics[scale=0.3]{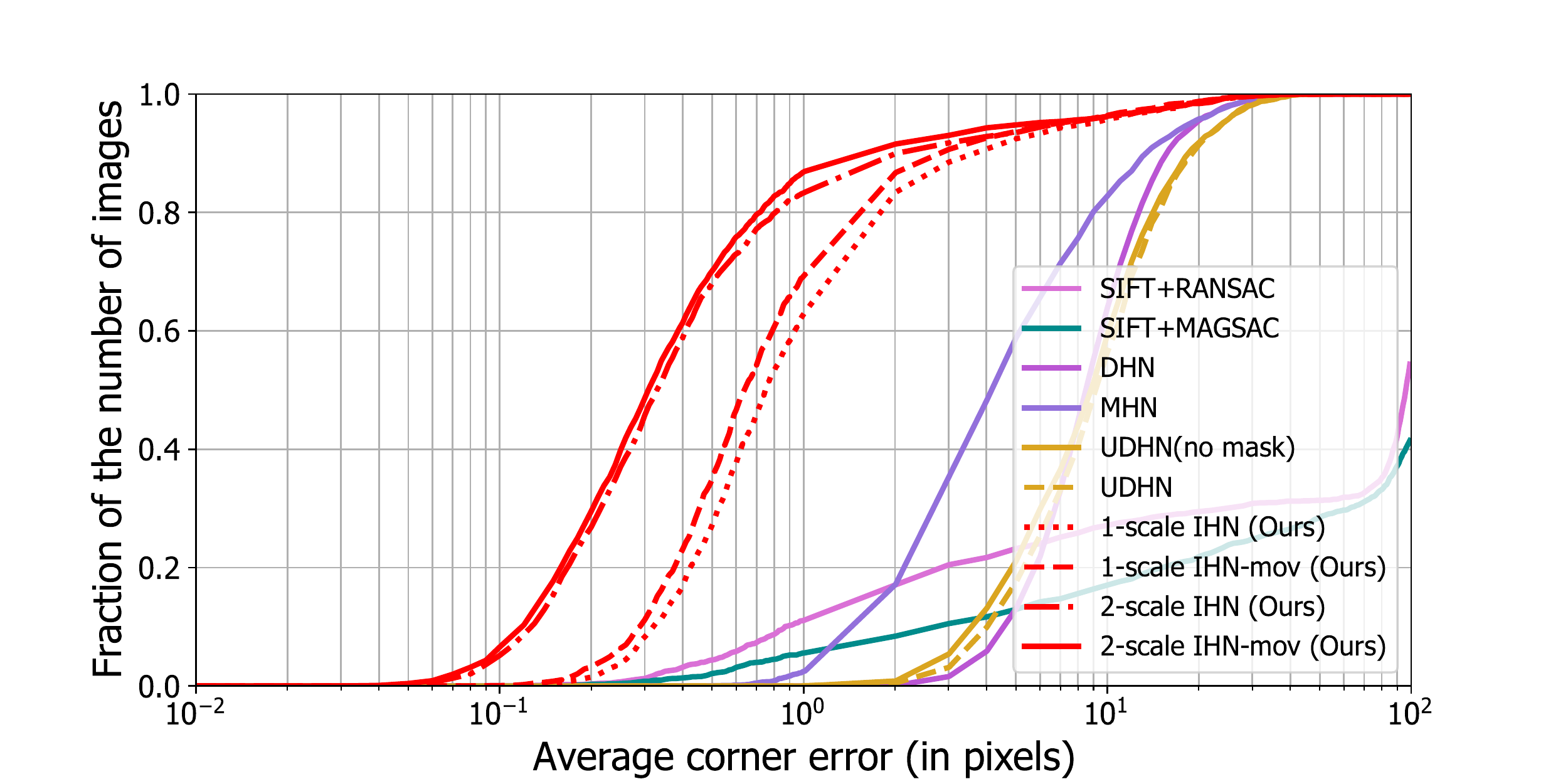}
  \vspace{-1mm}
  \caption{Evaluation on SPID.}
  \vspace{-1mm}
  \label{fig:eva_spid}
  \end{subfigure}
  \hfill
  \vspace{-2mm}
  \caption{Evaluation of homography estimation methods on MSCOCO, Google Earth, Google Map \& Satellite, and SPID datasets. MSCOCO contains common RGB images. Google Earth and Google Map \& Satellite are cross-modal datasets. SPID dataset provides surveillance images with foreground moving objects.} 
  \vspace{-3mm}
\end{figure*}
\section{Experiments}
\label{sec:experiments}

\subsection{Implementation Details}

We implement our network using PyTorch. The network is trained using AdamW \cite{loshchilov2017decoupled} optimizer and the maximum learning rate is set to $2.5 \times 10^{-4}$. The network is trained with a batch size of $16$ and a training iteration of $120000$. The total iteration $K$ at one resolution for training is set to $6$. We set the search radius of correlation updater $r = 4$ and set $\alpha=0.85$ in the loss function. We set $N=128$ for the GMA in the basic 1-scale IHN, and set $N=80$ for the GMA in the extra scale of the 2-scale IHN. For the mentioned different network structures and scales, we employ the same hyper-parameters.


\subsection{Datasets}

We evaluate IHN on both common and challenging datasets. We test IHN for static scenes on common MSCOCO dataset \cite{lin2014microsoft} as in \cite{detone2016deep,chang2017clkn,le2020deep,shao2021localtrans,zhao2021deep}. We also evaluate IHN on the cross-modal Google Earth and Google Map \& Satellite datasets \cite{zhao2021deep}. For dynamic scenes, we test IHN and IHN-mov on a challenging moving-objects scene generated based on the SPID surveillance dataset \cite{wang2016spid}.

Part of the images in SPID are discarded due to camera shake or overly low-quality. The images are processed in two different ways: (1) For scenes with low resolution, the original images are used directly. We randomly select two images within the same scene to make an image pair. (2) For scenes with high resolution, we randomly select one image first, and then use the provided pedestrian annotation to determine the position of foreground moving objects. We expand the pedestrian annotation to make the crop area contain both object and background. We then randomly select another image within the same scene, and crop at the same position to obtain an image pair that contains different foreground moving objects but the same background. We use 80\% images as the training data and 20\% the test data. 

For a fair comparison, all methods included for evaluation are trained and tested by the exact same corresponding training and test splits within each dataset.

\begin{table}\footnotesize
    \centering    
    \vspace{-3mm}
    \caption{Comparison of concatenation and iterative framework on MACE when using various network architectures.}
    \vspace{-3mm}
    \begin{tabular}{ccccc}
    \toprule
         & DHN \cite{detone2016deep} & MHN \cite{le2020deep} & UDHN \cite{zhang2020content} & GMA\\
    \midrule
    Concatenation  & 3.54 & 4.01 & 3.47 & 3.54\\

    Iterative framework      & 0.19 & 0.20 & 0.20 & 0.19\\
    \midrule
    Parameters     & 2.8 M & 2.2 M & 22.3 M & 0.8 M\\
    \bottomrule
    \end{tabular}
    \vspace{-6mm}
    \label{tab:iteration_ablation}
\end{table}
\subsection{Evaluation and Ablation Study on MSCOCO}

We evaluate our IHN on the MS-COCO dataset \cite{lin2014microsoft} with LocalTrans \cite{shao2021localtrans}, AffNet \cite{mishkin2018repeatability}, CLKN \cite{chang2017clkn}, LFNet \cite{ono2018lf}, DHN \cite{detone2016deep}, UDHN \cite{zhang2020content}, MHN \cite{le2020deep}, PFNet \cite{zeng2018rethinking}, PWC \cite{sun2018pwc}, SIFT+ContextDesc+RANSAC \cite{luo2019contextdesc}, SIFT+GeoDesc+RANSAC \cite{luo2018geodesc}, SIFT+MAGSAC \cite{barath2019magsac}, and SIFT+RANSAC \cite{lowe2004distinctive}. The experiment settings are the same as in most deep homography estimation methods \cite{detone2016deep,le2020deep,chang2017clkn,zhao2021deep}. The corners of $128 \times 128$ images are randomly shifted in the range of $[-32,32]$ pixels to make the deformed image. Similar to most deep homography works \cite{detone2016deep,chang2017clkn,le2020deep,shao2021localtrans,zhao2021deep}, we employ the average corner error (ACE) as the evaluation metric. An ablation study of our proposed iterative framework together with our network architecture is then conducted. All ablation study are conducted on the basic 1-scale IHN if not otherwise specifically mentioned.

\begin{table}\footnotesize
    \centering
    \vspace{-3mm}
    \caption{Ablation study on the settings of IHN.} 
    \vspace{-3mm}
    \begin{tabular}{cccc}
    \toprule
     Experiment & Setting & MACE &  Parameters \\
    \midrule
    \multirow{2}*{Correlation pooling}  & No pooling & 0.23 & 1.2 M \\
                                        & \textbf{Pooling} & {0.19} & 1.3 M \\
    \midrule
    \multirow{2}*{Parameterization}  & Homography & $\infty$ & 1.3 M \\
                                     & \textbf{Displacement} & {0.19} & 1.3 M \\
    \midrule
    \multirow{2}*{Homography flow}  & No flow & 0.21 & 1.3 M \\
                                     & \textbf{Flow} & {0.19} & 1.3 M \\
    \midrule
    \multirow{2}*{Scale}  & \textbf{$\mathbf{1}$ scale} & 0.19 & 1.3 M \\
                          & \textbf{$\mathbf{2}$ scales}& {0.06} & 1.7 M \\
    \midrule
    \multirow{4}*{Iteration}  & $1$ & 3.15 & 1.3 M \\
                          & $\mathbf{6}$  & {0.19} & 1.3 M \\
                          & $12$  & {0.19} & 1.3 M \\
                          & $100$  & {0.19} & 1.3 M \\
    \bottomrule
    \end{tabular}
    \vspace{-6mm}
    \label{tab:IHN_ablation}
\end{table}

\textbf{Evaluation on MSCOCO.} The statistical results on MSCOCO are illustrated in Fig. \ref{fig:eva_mscoco}. It is observed that the basic 1-scale IHN already outperforms most competitors except CLKN and LocalTrans. We note that MHN is conducted on $3$ scales, which is surpassed by our basic 1-scale IHN. Our 2-scale IHN outperforms all other homography estimation methods by a large gap. The 2-scale IHN produces over 90\% ACEs lower than $0.1$ pixels, which significantly outperforms CLKN that adopt the traditional IC-LK iterator.

\textbf{Ablation Study on the Iterative Framework.} To reveal the effectiveness of the iterative framework in IHE, we specifically conduct a homography estimation performance comparison on our iterative framework and the feature/image concatenation strategies in \cite{detone2016deep,le2020deep,zhang2020content,zhou2019stn,nguyen2018unsupervised}. We switch the global motion aggregator among the deep homography estimation architectures introduced in previous works including DHN\cite{detone2016deep}, MHN\cite{le2020deep}, UDHN\cite{zhang2020content}. To avoid the influence of the feature extractor, we uniformly use the feature extractor introduced in Section \ref{feature_extraction}. The extracted feature maps are concatenated in the channel dimension to achieve the feature/image concatenation strategy in \cite{detone2016deep,le2020deep,zhang2020content}. Table \ref{tab:iteration_ablation} lists the mean average corner error (MACE) and parameters of network using different architectures. It is observed that compared to the feature/image concatenation strategy, our iterative framework in IHE significantly boosts the estimation accuracy by about 95\% error reduction with the same network design. Also benefit from our iterative framework, our GMA can achieve comparable high accuracy with much fewer parameters compared to previous homography estimation architectures, \emph{e.g.}, 36.4\% of MHN \cite{le2020deep} and 3.6\% of UDHN \cite{zhang2020content}.

\textbf{Ablation Study on Settings of IHN.} Table \ref{tab:IHN_ablation} lists the mean average corner error (MACE) and parameters of IHN under different settings. The parameter counting involves the feature extractor introduced in Section \ref{feature_extraction}, which differs from Table \ref{tab:iteration_ablation}. The symbol $\infty$ denotes the case that the training doesn't converge and the MACE falls into infinity. It is observed that the correlation pooling, the parameterization using displacement, and the addition of homography flow improve the accuracy with very few parameter costs. The employment of another scale can improve the accuracy, despite that the 1-scale version already achieves relatively high accuracy. We further test the influence of inference iteration under the training iteration of $6$. It is observed that the network owns a significantly better performance at $6$ and $12$ iterations than $1$ iteration, once again indicating that iteration is vital for a high precision. Another interesting phenomenon is that when we raise the iteration to $100$, IHN is still stable and doesn't get into divergence. The bold setting options in Table \ref{tab:IHN_ablation} are employed in the rest experiments.

\subsection{Evaluation on Cross-Modal Datasets}
\label{subsec:cross-modal_exp}

We evaluate our IHN on cross-modal datasets \cite{zhao2021deep}, including Google Earth of season change and Google Map \& Satellite images with large modality difference. It is worth noting that we don't specifically alter IHN for cross-modal data as in \cite{zhao2021deep}, in which two separate feature extractors (namely pseudo-Siamese) are adopted. The original LK \cite{baker2004lucas}, SIFT+RANSAC\cite{lowe2004distinctive}, SIFT+MAGSAC\cite{barath2019magsac}, CLKN\cite{chang2017clkn}, DHN\cite{detone2016deep}, MHN\cite{le2020deep}, DHN+DLKFM\cite{zhao2021deep}, and MHN+DLKFM\cite{zhao2021deep} are included for comparison. As illustrated in Fig. \ref{fig:eva_ggearth} and \ref{fig:eva_ggmap}, our 1-scale and 2-scale IHN outperform the competitors by a large gap. We note that the recent MHN+DLKFM uses the combination of 3-scale VGG-style networks and 3-scales LK iterator. Our superiority over MHN+DLKFM further reveals the potential of the deep iterative framework. 

\begin{figure*}
  \center
  \begin{subfigure}{1\linewidth}
  \center
  \includegraphics[scale=0.34]{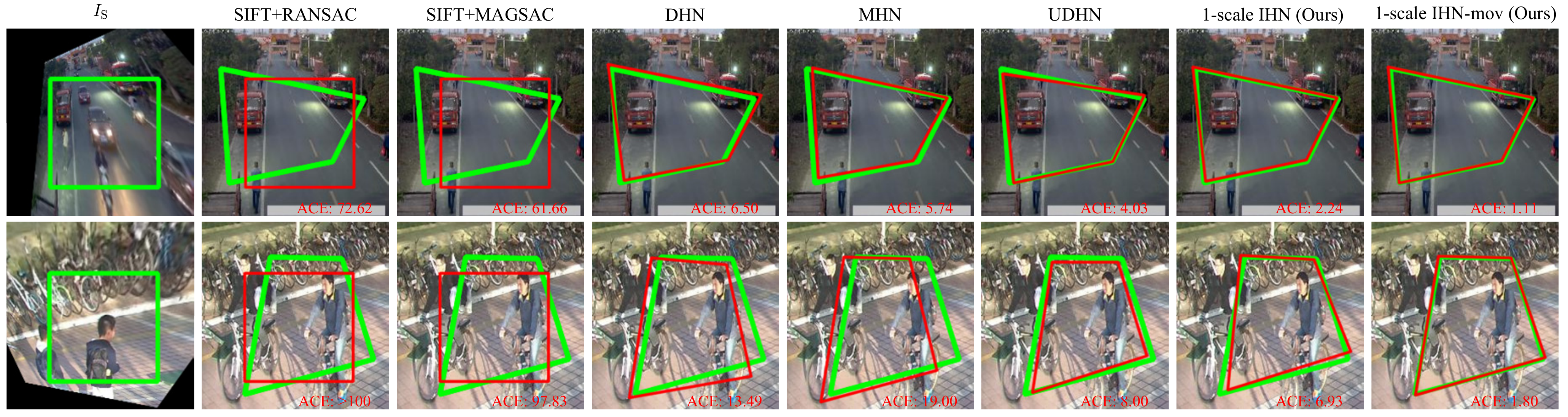}
  \vspace{-2mm}
    \caption{Visualization of homography estimation.}
    \label{fig:move_obj_ace}
  \end{subfigure}
  \hfill
  \centering
  \begin{subfigure}{1\linewidth}
  \center
  \includegraphics[scale=0.34]{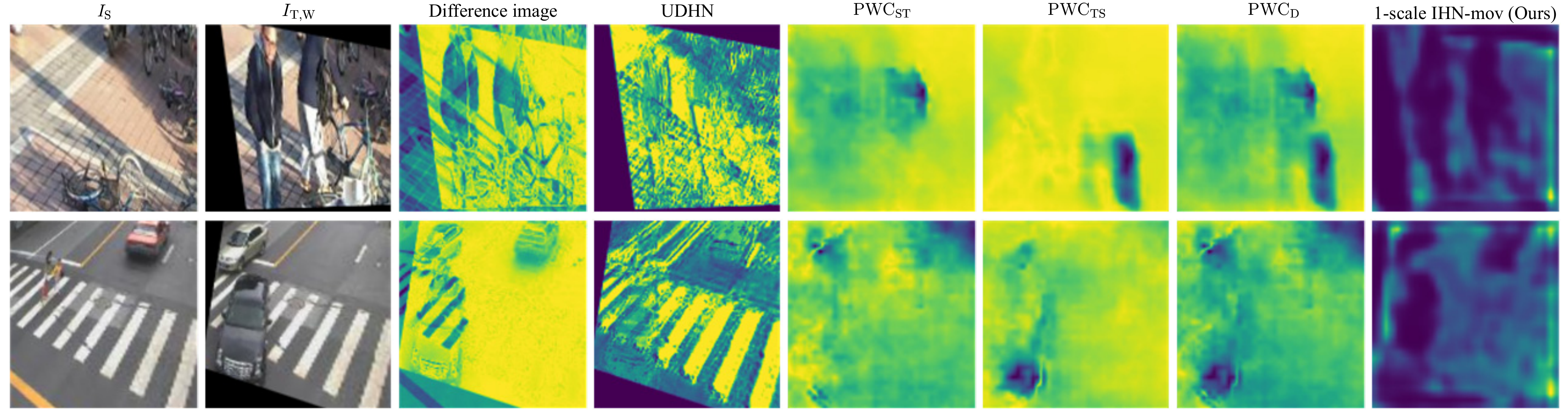}
   \vspace{-2mm}
    \caption{Visualization of inlier mask.}
  \label{fig:move_obj_mask}
  \end{subfigure}
   \vspace{-3mm}
  \caption{Visualization of results on the SPID dataset with moving objects. In (a), green polygons denote the ground-truth position of $I_\mathrm{S}$ on the target images. Red polygons denote the estimated position using different algorithms on the target images. The closer the 2 colors of polygons are, the better estimation accuracy  (also indicated by a lower ACE). In (b), the target images are warped to better illustrate the inlier masks. The mask produced by UDHN \cite{zhang2020content} detects the image content but not the needed matching inliers. The 3 masks produced by PWC-Net \cite{sun2018pwc} fail to predict the correct inliers. Our IHN-mov produces a more reasonable inlier mask by excluding the moving objects.}
   \vspace{-2mm}
\end{figure*}

\subsection{Evaluation on Moving-Objects Dataset}

We further conduct an evaluation on the SPID dataset that contains foreground moving objects. We compare our IHN with SIFT+RANSAC\cite{lowe2004distinctive}, SIFT+MAGSAC\cite{barath2019magsac}, DHN\cite{detone2016deep}, MHN\cite{le2020deep}, and UDHN\cite{zhang2020content}. The foreground moving objects usually occlude the background that satisfies the homography assumption, which makes accurate homography estimation difficult. We specifically propose an architecture for this scenario, called IHN-mov, which can produce a mask that explicitly weights the matching inliers to improve the estimation accuracy. We make UDHN into supervised for a fair comparison as its unsupervised training on SPID doesn't converge. As illustrated in Fig. \ref{fig:eva_spid}, 2 versions of IHN with different scales all outperform other competitors. 1-scale IHN-mov surpasses 1-scale IHN, with the fraction of ACE less than 1 pixel increased by around $6\%$. We further visualize the homography estimation of the above methods in Fig. \ref{fig:move_obj_ace}. It is observed that SIFT+RANSAC and SIFT+MAGSAC fail. The deep homography methods DHN, MHN, UDHN, and basic 1-scale IHN are affected by foreground moving objects, while 1-scale IHN-mov produces relatively more accurate homography estimations. The 2-scale IHN ourperforms 1-scale IHN-mov, meaning that the addition of extra scale can boost the estimation accuracy more. On the other side, the 2-scale IHN-mov exceeds 2-scale IHN, indicating that the weighting mask can boost the estimation within the same scale.

We further visualize the inlier mask produced by UDHN \cite{zhang2020content}, PWC-Net \cite{sun2018pwc} optical flow, and our 1-scale IHN-mov in Fig. \ref{fig:move_obj_mask}. As the ground-truth inlier mask is unavailable, we compute the difference of the source image $I_{\mathrm{S}}$ and the warped target image $I_{\mathrm{T,W}}$ and reverse its intensity to roughly illustrate the matching inliers. The mask of UDHN is obtained by dot product the separate UDHN masks of $I_{\mathrm{S}}$ and $I_{\mathrm{T,W}}$. We separately demonstrate the optical flow masks taking $I_{\mathrm{S}}$ as reference $\mathrm{PWC_{ST}}$, taking $I_{\mathrm{T}}$ as reference $\mathrm{PWC_{TS}}$ and their dot product $\mathrm{PWC_{D}}$. We reverse the normalized magnitude of the optical flow masks to show the weight of inliers. It is observed that the masks produced by UDHN are affected by image content. The shadow area in the left scene background is assigned with a very low weight, but it should belong to the matching inliers. As for the optical flow masks, the estimation fails as the moving objects vanish. On the contrary, our 1-scale IHN-mov produces the inlier masks directly from the local and global motion information like RANSAC, and thus is more reasonable. It's worth noting that we trained a version of UDHN without the weight mask, namely UDHN(no mask), the accuracy raises a little as shown in Fig. \ref{fig:eva_spid}.

\subsection{Cross Dataset Evaluation}
\label{subsec:cross-dataset_exp}

We conduct cross dataset evaluation of 1-scale IHN and DLKFM\cite{zhao2021deep} that uses untrainable IC-LK, with results listed in Table \ref{tab:IHN_cross}. It’s shown that the generalization ability of IHN is superior than the IC-LK based DLKFM. For IHN, the model trained on hard datasets (\eg Google Map \& Satellite ) generalizes better.

We evaluate our 1-scale IHN (trained on SPID) on a SfM dataset EPFL \cite{strecha2008benchmarking}. We include MHN\cite{le2020deep} and DPCP \cite{ding2020robust} (a method designed for estimating camera pose in the SfM pipeline). The mean errors of rotation and translation of camera pose are reported in Table \ref{tab:SfM}. It is observed that IHN owns the best accuracy for a good generalization ability.

\begin{table}\footnotesize
\renewcommand\arraystretch{1}
\renewcommand\tabcolsep{4pt}{}
    \centering
    \vspace{-3mm}
    \caption{Cross dataset evaluation of 1-scale IHN and DLKFM. Google Map means Google Map \& Satellite.} 
    \vspace{-3mm}
    \begin{tabular}{l|cc|cc|cc}
    \toprule
     \multirow{2}*{\diagbox[width=5em,height=2em,trim=l]{Train}{Test}} & \multicolumn{2}{c|}{MSCOCO} & \multicolumn{2}{c|}{Google Earth} & \multicolumn{2}{c}{Google Map} \\
     \cline{2-7}
     &\  IHN\  &DLKFM&\  IHN\  &DLKFM&\  IHN\  &DLKFM\\
  MSCOCO      &\textbf{0.19}&0.55&\textbf{2.41}&4.68&\textbf{24.21}&Fail\\
  Google Earth&\textbf{2.42}&10.16&\textbf{1.60}&3.88&\textbf{13.30}&39.85\\
  Google Map  &\textbf{1.72}&37.30&\textbf{3.06}&39.72&\textbf{0.92}&4.41\\

    \bottomrule
    \end{tabular}
    \vspace{-6mm}
    \label{tab:IHN_cross}
\end{table}

\begin{table}[b]\footnotesize
\renewcommand\arraystretch{0.95}
    \centering
    \vspace{-3mm}
    \caption{Cross dataset evaluation of 1-scale IHN, MHN, and DPCP on the EPFL SfM dataset.} 
    \vspace{-3mm}
    \begin{tabular}{lccc}
    \toprule

       & 1-scale IHN   & MHN & DPCP \\
    \midrule
      Rotation  &\textbf{4.38} & 19.54 & 5.44  \\
      Translation &\textbf{13.54}& 33.38 & 17.28 \\
    \bottomrule
    \end{tabular}
    \vspace{-3mm}
    \label{tab:SfM}
\end{table}

\subsection{Inference Time}

Table \ref{tab:time_cum} lists the inference time of 1-scale IHN, 2-scale IHN, 2-scale IHN-mov, conventional IC-LK iterator in DLKFM \cite{zhao2021deep}, and MHN+DLKFM. The comparison is conducted on an Intel Xeon Silver 4210R CPU @ 2.40GHz with 64GB memory and an NVIDIA Quadro RTX 8000. It is observed that the 1-scale IHN takes significantly less time. The speed of 1-scale IHN is 8$\times$ that of IC-LK iterator. Furthermore, 1-scale IHN is able to process image pairs in sequence at 32.7 fps. 

\begin{table}\footnotesize
\renewcommand\arraystretch{1}
\renewcommand\tabcolsep{3pt}{}
    \centering
    \vspace{-3mm}
    \caption{Inference time comparison (in milliseconds) of different types of IHN, DLKFM and MHN+DLKFM. "s." denotes scale.}
    \vspace{-3mm}
    \begin{tabular}{ccccc}
    \toprule
    1-s. IHN  & 2-s. IHN & 2-s. IHN-mov & DLKFM  &MHN+DLKFM  \\
    \midrule
      30.6  & 60.4 & 93.6 & 253.7 & 380.9 \\
    \bottomrule
    \end{tabular}    
    \vspace{-6mm}
    \label{tab:time_cum}
\end{table}

\section{Conclusions}

We have proposed a new iterative deep homography estimation architecture named IHN. Different from the iterative framework in the traditional IC-LK iterator, IHN is completely trainable. IHN achieves state-of-the-art performance on several challenging datasets, including cross-modal and moving-objects scenes. We experimentally show that the iterative framework of IHN is vital for the error reduction and parameter saving. IHN also has an advantage in speed over the traditional IC-LK algorithm. 

\textbf{Limitations. }The limitations of IHN are mainly 2 folds. First, the correlation volume raises the demand for GPU memory. Second, the resolution of the inlier mask produced by IHN-mov is limited by the feature map size and thus cannot be very clear. 

{\small
\bibliographystyle{ieee_fullname}
\bibliography{egbib}

\begin{thebibliography}{10}\itemsep=-1pt

\bibitem{abdel2015direct}
Yousset~I Abdel-Aziz, HM Karara, and Michael Hauck.
\newblock Direct linear transformation from comparator coordinates into object
  space coordinates in close-range photogrammetry.
\newblock {\em Photogrammetric Engineering \& Remote Sensing}, 81(2):103--107,
  2015.

\bibitem{baker2004lucas}
Simon Baker and Iain Matthews.
\newblock Lucas-{K}anade 20 years on: A unifying framework.
\newblock {\em International Journal of Computer Vision}, 56(3):221--255, 2004.

\bibitem{barath2019magsac}
Daniel Barath, Jiri Matas, and Jana Noskova.
\newblock {MAGSAC}: marginalizing sample consensus.
\newblock In {\em Proceedings of the IEEE/CVF Conference on Computer Vision and
  Pattern Recognition}, pages 10197--10205, 2019.

\bibitem{bay2006surf}
Herbert Bay, Tinne Tuytelaars, and Luc Van~Gool.
\newblock {SURF}: Speeded up robust features.
\newblock In {\em European Conference on Computer Vision}, pages 404--417.
  Springer, 2006.

\bibitem{chang2017clkn}
Che-Han Chang, Chun-Nan Chou, and Edward~Y Chang.
\newblock {CLKN}: Cascaded lucas-kanade networks for image alignment.
\newblock In {\em Proceedings of the IEEE Conference on Computer Vision and
  Pattern Recognition}, pages 2213--2221, 2017.

\bibitem{detone2016deep}
Daniel DeTone, Tomasz Malisiewicz, and Andrew Rabinovich.
\newblock Deep image homography estimation.
\newblock {\em arXiv preprint arXiv:1606.03798}, 2016.

\bibitem{ding2020robust}
Tianjiao Ding, Yunchen Yang, Zhihui Zhu, Daniel~P Robinson, Ren{\'e} Vidal,
  Laurent Kneip, and Manolis~C Tsakiris.
\newblock Robust homography estimation via dual principal component pursuit.
\newblock In {\em Proceedings of the IEEE/CVF Conference on Computer Vision and
  Pattern Recognition}, pages 6080--6089, 2020.

\bibitem{dubrofsky2009homography}
Elan Dubrofsky.
\newblock Homography estimation.
\newblock {\em Diplomov{\'a} pr{\'a}ce. Vancouver: Univerzita Britsk{\'e}
  Kolumbie}, 5, 2009.

\bibitem{engel2014lsd}
Jakob Engel, Thomas Sch{\"o}ps, and Daniel Cremers.
\newblock {LSD-SLAM}: Large-scale direct monocular slam.
\newblock In {\em European Conference on Computer Vision}, pages 834--849.
  Springer, 2014.

\bibitem{erlik2017homography}
Farzan Erlik~Nowruzi, Robert Laganiere, and Nathalie Japkowicz.
\newblock Homography estimation from image pairs with hierarchical
  convolutional networks.
\newblock In {\em Proceedings of the IEEE International Conference on Computer
  Vision Workshops}, pages 913--920, 2017.

\bibitem{fischler1981random}
Martin~A Fischler and Robert~C Bolles.
\newblock Random sample consensus: a paradigm for model fitting with
  applications to image analysis and automated cartography.
\newblock {\em Communications of the ACM}, 24(6):381--395, 1981.

\bibitem{goforth2019gps}
Hunter Goforth and Simon Lucey.
\newblock {GPS}-denied {UAV} localization using pre-existing satellite imagery.
\newblock In {\em 2019 International Conference on Robotics and Automation},
  pages 2974--2980. IEEE, 2019.

\bibitem{guo2016joint}
Heng Guo, Shuaicheng Liu, Tong He, Shuyuan Zhu, Bing Zeng, and Moncef Gabbouj.
\newblock Joint video stitching and stabilization from moving cameras.
\newblock {\em IEEE Transactions on Image Processing}, 25(11):5491--5503, 2016.

\bibitem{le2020deep}
Hoang Le, Feng Liu, Shu Zhang, and Aseem Agarwala.
\newblock Deep homography estimation for dynamic scenes.
\newblock In {\em Proceedings of the IEEE/CVF Conference on Computer Vision and
  Pattern Recognition}, pages 7652--7661, 2020.

\bibitem{lin2014microsoft}
Tsung-Yi Lin, Michael Maire, Serge Belongie, James Hays, Pietro Perona, Deva
  Ramanan, Piotr Doll{\'a}r, and C~Lawrence Zitnick.
\newblock Microsoft {COCO}: Common objects in context.
\newblock In {\em European Conference on Computer Vision}, pages 740--755.
  Springer, 2014.

\bibitem{liu2013bundled}
Shuaicheng Liu, Lu Yuan, Ping Tan, and Jian Sun.
\newblock Bundled camera paths for video stabilization.
\newblock {\em ACM Transactions on Graphics}, 32(4):1--10, 2013.

\bibitem{loshchilov2017decoupled}
Ilya Loshchilov and Frank Hutter.
\newblock Decoupled weight decay regularization.
\newblock {\em arXiv preprint arXiv:1711.05101}, 2017.

\bibitem{lowe2004distinctive}
David~G Lowe.
\newblock Distinctive image features from scale-invariant keypoints.
\newblock {\em International Journal of Computer Vision}, 60(2):91--110, 2004.

\bibitem{lucas1981iterative}
Bruce~D Lucas, Takeo Kanade, et~al.
\newblock An iterative image registration technique with an application to
  stereo vision.
\newblock In {\em Proceedings of the 7th International Joint Conference on
  Artificial intelligence}. Vancouver, British Columbia, 1981.

\bibitem{luo2019contextdesc}
Zixin Luo, Tianwei Shen, Lei Zhou, Jiahui Zhang, Yao Yao, Shiwei Li, Tian Fang,
  and Long Quan.
\newblock Contextdesc: Local descriptor augmentation with cross-modality
  context.
\newblock In {\em Proceedings of the IEEE/CVF Conference on Computer Vision and
  Pattern Recognition}, pages 2527--2536, 2019.

\bibitem{luo2018geodesc}
Zixin Luo, Tianwei Shen, Lei Zhou, Siyu Zhu, Runze Zhang, Yao Yao, Tian Fang,
  and Long Quan.
\newblock Geodesc: Learning local descriptors by integrating geometry
  constraints.
\newblock In {\em Proceedings of the European Conference on Computer Vision},
  pages 168--183, 2018.

\bibitem{maas2013rectifier}
Andrew~L Maas, Awni~Y Hannun, Andrew~Y Ng, et~al.
\newblock Rectifier nonlinearities improve neural network acoustic models.
\newblock In {\em Proceedings of International Conference on Machine Learning},
  volume~30, page~3. Citeseer, 2013.

\bibitem{mishkin2018repeatability}
Dmytro Mishkin, Filip Radenovic, and Jiri Matas.
\newblock Repeatability is not enough: Learning affine regions via
  discriminability.
\newblock In {\em Proceedings of the European Conference on Computer Vision},
  pages 284--300, 2018.

\bibitem{mur2015orb}
Raul Mur-Artal, Jose Maria~Martinez Montiel, and Juan~D Tardos.
\newblock {ORB-SLAM}: {A} versatile and accurate monocular slam system.
\newblock {\em IEEE Transactions on Robotics}, 31(5):1147--1163, 2015.

\bibitem{nguyen2018unsupervised}
Ty Nguyen, Steven~W Chen, Shreyas~S Shivakumar, Camillo~Jose Taylor, and Vijay
  Kumar.
\newblock Unsupervised deep homography: A fast and robust homography estimation
  model.
\newblock {\em IEEE Robotics and Automation Letters}, 3(3):2346--2353, 2018.

\bibitem{nocedal2006numerical}
Jorge Nocedal and Stephen Wright.
\newblock {\em Numerical optimization}.
\newblock Springer Science \& Business Media, 2006.

\bibitem{ono2018lf}
Yuki Ono, Eduard Trulls, Pascal Fua, and Kwang~Moo Yi.
\newblock {LF-N}et: Learning local features from images.
\newblock {\em arXiv preprint arXiv:1805.09662}, 2018.

\bibitem{shao2021localtrans}
Ruizhi Shao, Gaochang Wu, Yuemei Zhou, Ying Fu, Lu Fang, and Yebin Liu.
\newblock Localtrans: A multiscale local transformer network for
  cross-resolution homography estimation.
\newblock In {\em Proceedings of the IEEE/CVF International Conference on
  Computer Vision}, pages 14890--14899, 2021.

\bibitem{simon2000markerless}
Gilles Simon, Andrew~W Fitzgibbon, and Andrew Zisserman.
\newblock Markerless tracking using planar structures in the scene.
\newblock In {\em Proceedings IEEE and ACM International Symposium on Augmented
  Reality}, pages 120--128. IEEE, 2000.

\bibitem{strecha2008benchmarking}
Christoph Strecha, Wolfgang Von~Hansen, Luc Van~Gool, Pascal Fua, and Ulrich
  Thoennessen.
\newblock On benchmarking camera calibration and multi-view stereo for high
  resolution imagery.
\newblock In {\em Proceedings of the IEEE/CVF Conference on Computer Vision and
  Pattern Recognition}, pages 1--8. IEEE, 2008.

\bibitem{sun2018pwc}
Deqing Sun, Xiaodong Yang, Ming-Yu Liu, and Jan Kautz.
\newblock {PWC}-{N}et: {CNN}s for optical flow using pyramid, warping, and cost
  volume.
\newblock In {\em Proceedings of the IEEE Conference on Computer Vision and
  Pattern Recognition}, pages 8934--8943, 2018.

\bibitem{szeliski2006image}
Richard Szeliski.
\newblock Image alignment and stitching: A tutorial.
\newblock {\em Foundations and Trends{\textregistered} in Computer Graphics and
  Vision}, 2(1):1--104, 2006.

\bibitem{teed2020raft}
Zachary Teed and Jia Deng.
\newblock {RAFT}: Recurrent all-pairs field transforms for optical flow.
\newblock In {\em European Conference on Computer Vision}, pages 402--419.
  Springer, 2020.

\bibitem{wang2018deep}
Chaoyang Wang, Hamed~Kiani Galoogahi, Chen-Hsuan Lin, and Simon Lucey.
\newblock Deep-{LK} for efficient adaptive object tracking.
\newblock In {\em 2018 IEEE International Conference on Robotics and
  Automation}, pages 627--634. IEEE, 2018.

\bibitem{wang2016spid}
Dan Wang, Chongyang Zhang, Hao Cheng, Yanfeng Shang, and Lin Mei.
\newblock {SPID}: surveillance pedestrian image dataset and performance
  evaluation for pedestrian detection.
\newblock In {\em Asian Conference on Computer Vision}, pages 463--477.
  Springer, 2016.

\bibitem{wu2018group}
Yuxin Wu and Kaiming He.
\newblock Group normalization.
\newblock In {\em Proceedings of the European Conference on Computer Vision},
  pages 3--19, 2018.

\bibitem{ying2021unaligned}
Jiacheng Ying, Hui-Liang Shen, and Si-Yuan Cao.
\newblock Unaligned hyperspectral image fusion via registration and
  interpolation modeling.
\newblock {\em IEEE Transactions on Geoscience and Remote Sensing}, 2021.

\bibitem{zeng2018rethinking}
Rui Zeng, Simon Denman, Sridha Sridharan, and Clinton Fookes.
\newblock Rethinking planar homography estimation using perspective fields.
\newblock In {\em Asian Conference on Computer Vision}, pages 571--586.
  Springer, 2018.

\bibitem{zhang2020content}
Jirong Zhang, Chuan Wang, Shuaicheng Liu, Lanpeng Jia, Nianjin Ye, Jue Wang, Ji
  Zhou, and Jian Sun.
\newblock Content-aware unsupervised deep homography estimation.
\newblock In {\em European Conference on Computer Vision}, pages 653--669.
  Springer, 2020.

\bibitem{zhao2021deep}
Yiming Zhao, Xinming Huang, and Ziming Zhang.
\newblock Deep {L}ucas-{K}anade homography for multimodal image alignment.
\newblock In {\em Proceedings of the IEEE/CVF Conference on Computer Vision and
  Pattern Recognition}, pages 15950--15959, 2021.

\bibitem{zhou2019stn}
Qiang Zhou and Xin Li.
\newblock {STN}-homography: Direct estimation of homography parameters for
  image pairs.
\newblock {\em Applied Sciences}, 9(23):5187, 2019.

\bibitem{zhou2019integrated}
Yuan Zhou, Anand Rangarajan, and Paul~D Gader.
\newblock An integrated approach to registration and fusion of hyperspectral
  and multispectral images.
\newblock {\em IEEE Transactions on Geoscience and Remote Sensing},
  58(5):3020--3033, 2019.

\bibitem{zitova2003image}
Barbara Zitova and Jan Flusser.
\newblock Image registration methods: {A} survey.
\newblock {\em Image and Vision Computing}, 21(11):977--1000, 2003.

\end{thebibliography}
}

\end{document}